# Approximation Models of Combat in StarCraft 2

Ian Helmke, Daniel Kreymer, and Karl Wiegand
Northeastern University
Boston, MA 02115
{ihelmke, dkreymer, wiegandkarl} @gmail.com
December 3, 2012

## Abstract

Real-time strategy (RTS) games make heavy use of artificial intelligence (AI), especially in the design of computerized opponents. Because of the computational complexity involved in managing all aspects of these games, many AI opponents are designed to optimize only a few areas of playing style. In games like StarCraft 2, a very popular and recently released RTS, most AI strategies revolve around economic and building efficiency: AI opponents try to gather and spend all resources as quickly and effectively as possible while ensuring that no units are idle. The aim of this work was to help address the need for AI combat strategies that are not computationally intensive. Our goal was to produce a computationally efficient model that is accurate at predicting the results of complex battles between diverse armies, including which army will win and how many units will remain. Our results suggest it may be possible to develop a relatively simple approximation model of combat that can accurately predict many battles that do not involve micromanagement. Future designs of AI opponents may be able to incorporate such an approximation model into their decision and planning systems to provide a challenge that is strategically balanced across all aspects of play.

## Background and Motivation

StarCraft 2 is a real-time strategy (RTS) computer game created by Blizzard Entertainment. This game is a sequel to the original StarCraft and, like its predecessor, has proven very popular. Released in July 2010, StarCraft 2 sold over 3 million copies worldwide after only one month on the market, making it one of the fastest-selling RTS games of all time.[1] Many professional computer gaming leagues that previously ran StarCraft tournaments have recently converted to running StarCraft 2 tournaments, including the GOMTV Global StarCraft 2 League (GSL), which has awarded approximately $2.5 million USD in prize money since its inception in 2010.[2] The rising popularity of "e-sports" and the potential of winning substantial prize money in tournaments have contributed to an increased interest in strategies and technical analyses of many competitive computer games, especially StarCraft 2.[3]

StarCraft 2, like many other titles in the RTS genre, is a game for two or more players. Each player chooses from one of three different races (Protoss, Zerg, or Human) and builds an army of units to defeat his or her enemies on a specific map: a 2-dimensional grid of finite size with various terrain artifacts such as resources and obstructions. To build their armies, players must mine resources, construct buildings, research technology, and train units, all in real time. Additionally, players must balance their time and efforts between creating a strong economy, which enables faster production of units as well as the production of more advanced units, and attacking their opponents. Players win a game by destroying all of their opponents' buildings. Although there is no limit on the length of a game, many games last between 5 and 30 minutes.

Each of the three races in StarCraft 2 has its own unique behaviors, buildings, units, and technologies that allow for many different styles of play. Each map in StarCraft 2 is also unique, allowing for different strategies that take advantage of the terrain and positioning of resources. Because of the open-ended nature of the game, the computational complexity involved in creating artificially intelligent (AI) opponents is very high. AI opponents are typically built to simply brute-force efficiency, issuing many hundreds of orders per minute with the goal of overwhelming human opponents who cannot physically or mentally match the micromanagement capabilities of a computer.

These AI opponents, however, are often still defeated by human players through macromanagement strategies that focus on technologies, positioning, elements of surprise, and weaknesses in AI behavior (e.g. suboptimal reactions to being attacked very early or being attacked from the rear). Experienced players may exploit build timing such that they are able to attack just as they achieve a key technological or army advantage over their opponent. Additionally, humans are still better than AI players at some types of micromanagement. For example, while an AI player may be able to micromanage more units in the same period of time, humans may be more skilled at determining which units to micromanage, and how.

## Prior Work

StarCraft 2 includes a stock AI with several difficulty settings; however, these AIs are fairly simplistic and make no use of advanced micromanagement techniques or sophisticated decision-making; they are primarily intended to help new players learn how to play the game. The highest levels of StarCraft 2 AI actually cheat. The "insane" AI, for example, has special workers that return 40% more minerals per trip than any other workers.

Some attempts at a StarCraft 2 AI build on the framework provided by the default StarCraft 2 AIs, but still cheat. The Green Tea AI is one such attempt: it implements more sophisticated tactics than the Blizzard StarCraft AI, but cheats through revealing the map, so that it receives perfect information, while the player is still subject to the fog of war. On higher difficulties, this AI also gathers resources faster than it should, similar to the "insane" Blizzard AI.[4] This AI also implements "personalities" that use different high-level tactics: some try to expand frequently, some are more aggressive, etc.

There have been several attempts at making more sophisticated AIs. One such attempt by Matt Fisher intercepts DirectX API calls from the StarCraft 2 game in order to build a picture of the game using image processing, determining information about the game state based on the information actually shown to a player on screen.[5] The Fisher AI does not cheat.

To our knowledge there has been no formal work on building a model to predict the outcome of engagements between groups of units in StarCraft 2. There has been at least one attempt to create a crowdsourced battle simulator, in which players can input requests for different armies to battle each other, and the simulator queues and executes them.[6] However, this simulator appears defunct and does not actually use a combat model; instead, it simply executes battles within the Starcraft 2 Engine. One iPhone application appears to perform some kind of simulation given two groups of units, though the specifics are unknown.[7] There are also unit-vs-unit comparisons available that detail the number of attacks one unit needs to destroy another.[8] This is useful as a unit reference but provides no direct insight into the behavior of armies and the outcome of complex engagements.

## Approach

The current work aimed to investigate different approximation models for StarCraft 2, specifically for combat between groups of different units. Our hypothesis was that relatively little information about the composition of an army is necessary in order to accurately model combat. To test our hypothesis, we designed and implemented several models that could allow an AI opponent to determine whether an engagement will end favorably, at as low a computational cost as possible. We quantified each model's accuracy against in-game test results as the size and composition of the engaging armies was varied.

### Approximating Battles

We designed and developed several different approximation models for combat. Each model builds upon the previous model in an additive fashion. Additionally, all of our approximation models assume:

1. Two opposing armies;
2. Each army begins just out of sight, at the edge of the fog-of-war;
3. Each army is grouped together in a relatively tight, random cluster;
4. Each army receives an "attack-move" command past the opposing army at relatively the same time; and,
5. No army is micromanaged.

While many battles in StarCraft 2, especially at the professional level, involve intense micromanagement, many successful players also use macromanagement strategies. We chose to begin with approximations designed to emulate the behavior of battles without micromanagement, with the hope that micromanagement strategies could be added later.

> *APX1*
> Our base approximation model simulates the effects of one-second "rounds" by randomly selecting units from each army and applying damage until only one army remains. From a practical standpoint, this is similar to the concept of simultaneous "perfect focus fire" with randomly selected targets. The algorithm for APX1 is as follows:

```
getAPX1 (army1, army2):
    while (getHealth(army1) > 0 and getHealth(army2) > 0):
        dam1 = getDPS(army1)
        dam2 = getDPS(army2)
        while (dam1 > 0):
            unit = getRandomUnit(army2)
            dam1 -= unit.health
            probabilisticallyKillUnit(unit, dam1)
        while (dam2 > 0):
            unit = getRandomUnit(army2)
            dam1 -= unit.health
            probabilisticallyKillUnit(unit, dam2)
    return (army1, army2)
```

*APX2*

Our second approximation model builds upon the base model by accounting for range. We chose to model range as free damage, simulating attacks that ranged units get before they are within melee distance and can be counter-attacked. The algorithm for APX2 is as follows:

```
getAPX2 (army1, army2):
    round1 = True
    while (getHealth(army1) > 0 and getHealth(army2) > 0):
        if round1:
            dam1 = getRangedDPS(army1)
            dam2 = getRangedDPS(army2)
            round1 = False
        else:
            dam1 = getDPS(army1)
            dam2 = getDPS(army2)
        while (dam1 > 0):
            killRandomUnit(army2)
        while (dam2 > 0):
            killRandomUnit(army1)
    return (army1, army2)
```

*APX3*

Our third approximation model adds the concept of bonus damage: some units in StarCraft 2 do more damage to units with certain attributes. We chose to model bonus damage as a single pool based on the percentage of units in the opposing army with the vulnerable attribute. So, a group of Hellions, which do bonus damage against units with the Light attribute, would add 50% of their potential bonus damage to their army's DPS calculation if 50% of the units in the opposing army have the Light attribute. The algorithm for APX3 is as follows:

```
getAPX3 (army1, army2):
    round1 = True
    while (getHealth(army1) > 0 and getHealth(army2) > 0):
        if round1:
            dam1 = getRangedDPS(army1) +
                    getRangedBonusDPS(army1, army2)
            dam2 = getRangedDPS(army2) +
                    getRangedBonusDPS(army2, army1)
            round1 = False
        else:
            dam1 = getDPS(army1) + getBonusDPS(army1, army2)
            dam2 = getDPS(army2) + getBonusDPS(army2, army1)
        while (dam1 > 0):
            killRandomUnit(army2)
        while (dam2 > 0):
            killRandomUnit(army1)
    return (army1, army2)
```

*APX4*

Our fourth approximation model prioritizes melee units over ranged units. The unit AI in StarCraft 2 targets the closest enemy unit if a specific unit is not selected. In practice, this means that melee units generally get attacked first. The only modification for this model is that instead of killing a random unit, we kill a random melee unit until none exist, and then we proceed to kill random ranged units. The algorithm for APX4 is:

```
getAPX4 (army1, army2):
    round1 = True
    while (getHealth(army1) > 0 and getHealth(army2) > 0):
        if round1:
            dam1 = getRangedDPS(army1) +
                    getRangedBonusDPS(army1, army2)
            dam2 = getRangedDPS(army2) +
                    getRangedBonusDPS(army2, army1)
            round1 = False
        else:
            dam1 = getDPS(army1) + getBonusDPS(army1, army2)
```

```
                dam2 = getDPS(army2) + getBonusDPS(army2,
        army1)
        while (dam1 > 0):
                killRandomUnit(army2, preferMelee = True)
        while (dam2 > 0):
                killRandomUnit(army1, preferMelee = True)
        return (army1, army2)
```

**Approximating Units**

Because battles in StarCraft 2 involve units, any approximation model of combat must also approximate units and unit behavior. To this end, we designed a base representation of a unit as having the following attributes:

1. **Health**: An integer number representing the starting amount of health for this unit plus the starting amount of shields, if applicable (e.g. for Zealots, this was 100 health + 50 shields = 150 health).
2. **DPS**: A decimal number representing the amount of damage done per second by this unit; for units with an area-of-effect attack (e.g. Hellions), the DPS was equal to the base DPS multiplied by the maximum potential area-of-effect (e.g. for Hellions, this was 3.2 DPS * 5 area = 16 DPS). Our reasoning was that in a sufficiently large engagement, a unit with an area-of-effect attack will usually be able to target multiple units effectively.
3. **Ranged**: A boolean value indicating whether the unit's attack is a ranged attack.
4. **Armor**: An integer number representing the amount of armor this unit has; this number was used to add a multiplicative amount of health (1.5) to the unit to simulate armor (e.g. for Zealots, this was 1.5 * 1 armor * 150 health = 225 health).
5. **Attributes**: A list of string attributes of the unit, such as Biological, Light, or Armored.
6. **Bonuses**: A list of string attributes that indicate a targeted unit will receive more damage from this unit.
7. **Bonus DPS**: A decimal number representing the amount of bonus damage done per second by this unit; for units with an area-of-effect attack (e.g. Hellions), the Bonus DPS was equal to the base bonus DPS multiplied by the maximum potential area-of-effect (e.g. for Hellions, this was 2.4 DPS * 5 area = 12 DPS).

## Method

We built 12 customized StarCraft 2 maps, each representing a match between two opposing armies of different races. For each match, we ran 10 battles within the StarCraft 2 engine and recorded the results, including how many times each army won and how many units remained for each army when it won. For each match, the remaining army compositions were averaged over the 10 battles to produce an overall result. These experimental results were compared with the results from our simulation engine, which ran 1000 rounds of each approximation model.

Our 12 matches were organized into 4 rounds of increasing army complexity:

**Round 1**
1. PvT: 8 Zealots and 2 Stalkers vs. 12 Marines and 4 Marauders
2. TvZ: 12 Marines and 4 Marauders vs. 24 Zerglings and 4 Roaches
3. PvZ: 8 Zealots and 2 Stalkers vs. 24 Zerglings and 4 Roaches

**Round 2**
1. PvT: 8 Zealots, 2 Stalkers, and 2 Sentries vs. 12 Marines, 4 Marauders, and 4 Hellions
2. TvZ: 12 Marines, 4 Marauders, and 4 Hellions vs. 24 Zerglings, 4 Roaches, and 4 Hydralisks
3. PvZ: 8 Zealots, 2 Stalkers, and 2 Sentries vs. 24 Zerglings, 4 Roaches, and 4 Hydralisks

**Round 3**
1. PvT: 8 Zealots, 2 Stalkers, 2 Sentries, and 1 Archon vs. 12 Marines, 4 Marauders, 4 Hellions, and 2 Tanks
2. TvZ: 12 Marines, 4 Marauders, 4 Hellions, and 2 Tanks vs. 24 Zerglings, 4 Roaches, 4 Hydralisks, and 1 Ultralisk
3. PvZ: 8 Zealots, 2 Stalkers, 2 Sentries, and 1 Archon vs. 24 Zerglings, 4 Roaches, 4 Hydralisks, and 1 Ultralisk

**Round 4**
1. PvT: 20 Zealots, 14 Stalkers, 2 Immortals, and 2 Colossi vs. 30 Marines, 10 Marauders, 4 Tanks, and 2 Thors
2. TvZ: 30 Marines, 10 Marauders, 4 tanks, and 2 Thors vs. 40 Zerglings, 10 Roaches, 10 Hydralisks, and 4 Ultralisks
3. PvZ: 20 Zealots, 14 Stalkers, 2 Immortals, and 2 Colossi vs. 40 Zerglings, 10 Roaches, 10 Hydralisks, and 4 Ultralisks

## Results

Table 1 shows a comparison of the experimental results for each match compared to the predictions of each of our approximation models. A value of "Test" in the Type column indicates that the row summarizes experimental results of the match. A value of "XvY" in the Match column indicates a battle between race X and race Y, where each race is one of Protoss, Terran, or Zerg (i.e. "TvZ" means a match between Terran and Zerg). Each column of the form "1 - #" indicates a unit type in army 1; each column of the form "2 - #" indicates a unit type in army 2. For comparison purposes, each unit type in these columns is consistent across each race and each column number represents a unit of the same general caliber (i.e. similar technology level, health, damage, or resource costs). The "1 - %" and "2 - %" columns indicate the percentage of battles won by army 1 and army 2, respectively; these percentages add to 1 in every row.

Our results show that in Round 1, our approximation models are able to accurately predict the winning army in the Protoss-Terran (81% Terran wins) conflict, but are less accurate at predicting both the Terran-Zerg matches (55% Terran wins versus an actual 70%) and the Protoss-Zerg matches (66% Protoss wins versus an actual 50%). Additionally, we see that the accuracy, or inaccuracy, of our approximation models does not drastically change when considering range (APX2) or bonus damage (APX3), but is moderately improved when preferring melee units over ranged units (APX4). Although both the Protoss-Zerg and Protoss-Terran matches become slightly less accurate in APX4, the Terran-Zerg match becomes much more accurate, moving from 7% to 55% with a goal of 70%.

For Round 2 matches, our approximation models are able to accurately predict both the winning army and the remaining army compositions, although the win percentage is overestimated in the Terran-Zerg match (88% Terran wins versus an actual 50%). In the Round 2 simulations, we can also see the effects of both range and bonus damage on estimation accuracy. For the Protoss-Terran match, our experimental results show that the winning army is always Terran (100%). Our APX1 model, however, estimates the Terran winning with 17% likelihood. Adding range (APX2) brings the Terran estimate to 62%, adding bonus damage (APX3) it to 89%, and preferring melee targets brings the final estimate to 94%. We see similar behavior in the Protoss-Zerg match, where APX2 and APX3 gradually bring the odds of either army winning closer to 50-50, and APX4 makes the remaining army composition more accurate at the expense of overestimating Terran victory.

In Round 3, we again see problems with the Terran-Zerg match: APX4 estimates the Terran army's chances at approximately 19% when test results show a 90% chance. APX4 also lowers the accuracy of the Protoss-Terran match and the Protoss-Zerg match compared to APX3. The Protoss-Terran match in Round 3 provides more evidence of the importance in considering both range and bonus damage as we observe the winning Protoss percentage go from 90% in APX1 to 64% in APX2 and 56% in APX3, compared to the 10% obtained from our tests. The preference for targeting melee units in APX4 reversed this improvement.

In Round 4, we simulate very large battles. Here, our model remains accurate in the Protoss-Terran and Protoss-Zerg matches, with both showing clear Protoss wins (100% and 77% compared to 100%). APX3 is less accurate than APX2 in the Protoss-Zerg match (from 70% to 50% with a goal of 100%), but APX4 brings the final estimate to 77% chance of a Protoss victory. Finally, our model proves to be entirely inaccurate in the Terran-Zerg case, indicating a clear Zerg win where in trials Zerg only won 40% of the time.

| Round | Type | Match | 1-1 | 1-2 | 1-3 | 1-4 | 2-1 | 2-2 | 2-3 | 2-4 | 1-% | 2-% |
|---|---|---|---|---|---|---|---|---|---|---|---|---|
| 1 | Test | PvT | 4 | 1 | 0 | 0 | 3 | 0 | 0 | 0 | 0.92 | 0.08 |
| 1 | APX1 | PvT | 6 | 1 | 0 | 0 | 0 | 0 | 0 | 0 | 0.99 | 0.01 |
| 1 | APX2 | PvT | 4 | 1 | 0 | 0 | 0 | 0 | 0 | 0 | 0.93 | 0.07 |
| 1 | APX3 | PvT | 4 | 1 | 0 | 0 | 0 | 0 | 0 | 0 | 0.92 | 0.08 |
| 1 | APX4 | PvT | 3 | 2 | 0 | 0 | 0 | 0 | 0 | 0 | 0.81 | 0.19 |
| 1 | Test | TvZ | 5 | 2 | 0 | 0 | 0 | 2 | 0 | 0 | 0.70 | 0.30 |
| 1 | APX1 | TvZ | 0 | 0 | 0 | 0 | 14 | 3 | 0 | 0 | 0.00 | 1.00 |
| 1 | APX2 | TvZ | 0 | 0 | 0 | 0 | 9 | 3 | 0 | 0 | 0.03 | 0.97 |
| 1 | APX3 | TvZ | 0 | 0 | 0 | 0 | 8 | 2 | 0 | 0 | 0.07 | 0.93 |
| 1 | APX4 | TvZ | 1 | 1 | 0 | 0 | 0 | 1 | 0 | 0 | 0.55 | 0.45 |
| 1 | Test | PvZ | 4 | 2 | 0 | 0 | 0 | 2 | 0 | 0 | 0.50 | 0.50 |
| 1 | APX1 | PvZ | 1 | 0 | 0 | 0 | 3 | 2 | 0 | 0 | 0.43 | 0.57 |
| 1 | APX2 | PvZ | 1 | 0 | 0 | 0 | 3 | 2 | 0 | 0 | 0.41 | 0.59 |
| 1 | APX3 | PvZ | 1 | 0 | 0 | 0 | 3 | 2 | 0 | 0 | 0.40 | 0.60 |
| 1 | APX4 | PvZ | 2 | 1 | 0 | 0 | 1 | 1 | 0 | 0 | 0.66 | 0.35 |
| 2 | Test | PvT | 0 | 0 | 0 | 0 | 4 | 4 | 1 | 0 | 0.00 | 1.00 |
| 2 | APX1 | PvT | 4 | 1 | 0 | 0 | 0 | 0 | 0 | 0 | 0.83 | 0.17 |
| 2 | APX2 | PvT | 2 | 1 | 0 | 0 | 1 | 1 | 0 | 0 | 0.38 | 0.62 |
| 2 | APX3 | PvT | 0 | 0 | 0 | 0 | 4 | 2 | 2 | 0 | 0.11 | 0.89 |
| 2 | APX4 | PvT | 0 | 0 | 0 | 0 | 5 | 2 | 3 | 0 | 0.06 | 0.94 |
| 2 | Test | TvZ | 2 | 2 | 1 | 0 | 0 | 0 | 3 | 0 | 0.50 | 0.50 |
| 2 | APX1 | TvZ | 0 | 0 | 0 | 0 | 12 | 3 | 2 | 0 | 0.00 | 1.00 |
| 2 | APX2 | TvZ | 0 | 0 | 0 | 0 | 6 | 2 | 1 | 0 | 0.13 | 0.87 |
| 2 | APX3 | TvZ | 1 | 1 | 1 | 0 | 3 | 1 | 1 | 0 | 0.43 | 0.57 |
| 2 | APX4 | TvZ | 3 | 2 | 2 | 0 | 0 | 0 | 0 | 0 | 0.88 | 0.12 |
| 2 | Test | PvZ | 0 | 0 | 0 | 0 | 0 | 4 | 4 | 0 | 0.00 | 1.00 |
| 2 | APX1 | PvZ | 0 | 0 | 0 | 0 | 8 | 3 | 2 | 0 | 0.06 | 0.94 |
| 2 | APX2 | PvZ | 0 | 0 | 0 | 0 | 9 | 3 | 2 | 0 | 0.03 | 0.97 |
| 2 | APX3 | PvZ | 0 | 0 | 0 | 0 | 9 | 3 | 2 | 0 | 0.02 | 0.98 |
| 2 | APX4 | PvZ | 0 | 0 | 0 | 0 | 2 | 4 | 3 | 0 | 0.04 | 0.96 |
| 3 | Test | PvT | 0 | 1 | 0 | 0 | 1 | 2 | 1 | 2 | 0.10 | 0.90 |
| 3 | APX1 | PvT | 2 | 1 | 0 | 0 | 1 | 0 | 0 | 0 | 0.90 | 0.10 |
| 3 | APX2 | PvT | 1 | 0 | 0 | 0 | 2 | 1 | 1 | 1 | 0.64 | 0.36 |
| 3 | APX3 | PvT | 0 | 0 | 0 | 0 | 4 | 1 | 2 | 1 | 0.56 | 0.44 |
| 3 | APX4 | PvT | 0 | 1 | 1 | 1 | 0 | 0 | 0 | 0 | 0.79 | 0.21 |
| 3 | Test | TvZ | 1 | 3 | 0 | 2 | 0 | 0 | 4 | 0 | 0.90 | 0.10 |
| 3 | APX1 | TvZ | 0 | 0 | 0 | 0 | 8 | 3 | 2 | 1 | 0.00 | 1.00 |
| 3 | APX2 | TvZ | 0 | 0 | 0 | 0 | 4 | 1 | 1 | 1 | 0.03 | 0.97 |
| 3 | APX3 | TvZ | 1 | 1 | 0 | 0 | 2 | 1 | 1 | 0 | 0.07 | 0.93 |
| 3 | APX4 | TvZ | 0 | 0 | 0 | 0 | 0 | 1 | 1 | 2 | 0.19 | 0.81 |
| 3 | Test | PvZ | 0 | 0 | 0 | 0 | 0 | 4 | 4 | 1 | 0.00 | 1.00 |
| 3 | APX1 | PvZ | 0 | 0 | 0 | 0 | 7 | 3 | 2 | 1 | 0.06 | 0.94 |
| 3 | APX2 | PvZ | 0 | 0 | 0 | 0 | 8 | 3 | 2 | 1 | 0.04 | 0.96 |
| 3 | APX3 | PvZ | 0 | 0 | 0 | 0 | 7 | 3 | 2 | 1 | 0.19 | 0.81 |
| 3 | APX4 | PvZ | 0 | 1 | 0 | 0 | 0 | 1 | 1 | 1 | 0.44 | 0.56 |
| 4 | Test | PvT | 1 | 12 | 2 | 2 | 0 | 0 | 0 | 0 | 1.00 | 0.00 |
| 4 | APX1 | PvT | 15 | 11 | 2 | 2 | 0 | 0 | 0 | 0 | 1.00 | 0.00 |
| 4 | APX2 | PvT | 14 | 10 | 2 | 2 | 0 | 0 | 0 | 0 | 1.00 | 0.00 |
| 4 | APX3 | PvT | 13 | 10 | 2 | 2 | 0 | 0 | 0 | 0 | 1.00 | 0.00 |
| 4 | APX4 | PvT | 7 | 14 | 2 | 2 | 0 | 0 | 0 | 0 | 1.00 | 0.00 |
| 4 | Test | TvZ | 0 | 3 | 2 | 0 | 0 | 1 | 7 | 0 | 0.60 | 0.40 |
| 4 | APX1 | TvZ | 0 | 0 | 0 | 0 | 24 | 7 | 6 | 4 | 0.00 | 1.00 |
| 4 | APX2 | TvZ | 0 | 0 | 0 | 0 | 24 | 6 | 6 | 3 | 0.00 | 1.00 |
| 4 | APX3 | TvZ | 0 | 0 | 0 | 0 | 24 | 6 | 6 | 3 | 0.00 | 1.00 |
| 4 | APX4 | TvZ | 0 | 0 | 0 | 0 | 0 | 7 | 6 | 4 | 0.00 | 1.00 |
| 4 | Test | PvZ | 0 | 8 | 2 | 2 | 0 | 0 | 0 | 0 | 1.00 | 0.00 |
| 4 | APX1 | PvZ | 5 | 3 | 1 | 1 | 1 | 0 | 0 | 1 | 0.75 | 0.26 |
| 4 | APX2 | PvZ | 4 | 3 | 1 | 1 | 1 | 1 | 0 | 1 | 0.70 | 0.30 |
| 4 | APX3 | PvZ | 3 | 2 | 0 | 1 | 2 | 1 | 1 | 1 | 0.55 | 0.45 |
| 4 | APX4 | PvZ | 0 | 6 | 1 | 1 | 0 | 0 | 0 | 0 | 0.77 | 0.23 |

Table 1: Summarized Experimental Results of Each Match and Simulation

## Discussion

Range provided a large improvement in nearly every engagement our original model had trouble predicting. Specifically, the Round 2 Protoss-Terran simulations went from inaccurately predicting a Protoss victory with 83% probability to predicting a Protoss victory with only 23% probability (Terran won the match in a simulated engagement 100% of the time). Several other matches saw similar gains. Almost all Terran units are ranged, and thus the improvements in APX2 greatly improved our accuracy where Terran was the favored army.

We also saw gains when we applied unit damage bonuses to both armies, as we see in the changes in our model outcomes from APX2 to APX3. One pronounced accuracy gain from APX2 to APX3 is in our Round 2 Terran-Zerg match. Both Marauders and Hellions have bonus DPS against certain types of units, and both of those types of units are found within the Zerg army. As a result, Terran saw a large DPS boost in this match, though this was not sufficient to render the model accurate (43% in APX3 and 88% in APX4 compared to a 100% win rate in trials).

The impact of targeting priority is less clear. Targeting melee units was a large gain in the round 2 Terran-Zerg match: the Terran win rate was 88% in our APX4 model compared to 43% at APX3 (Terran had a favorable outcome 100% of the time in trial runs). On the other hand, our Protoss-Zerg Round 3 match prediction became worse relative to our trial runs. This may be due in part to the fact that Ultralisks are difficult to destroy with their high health and armor, resulting in the Protoss army requiring multiple rounds to destroy one.

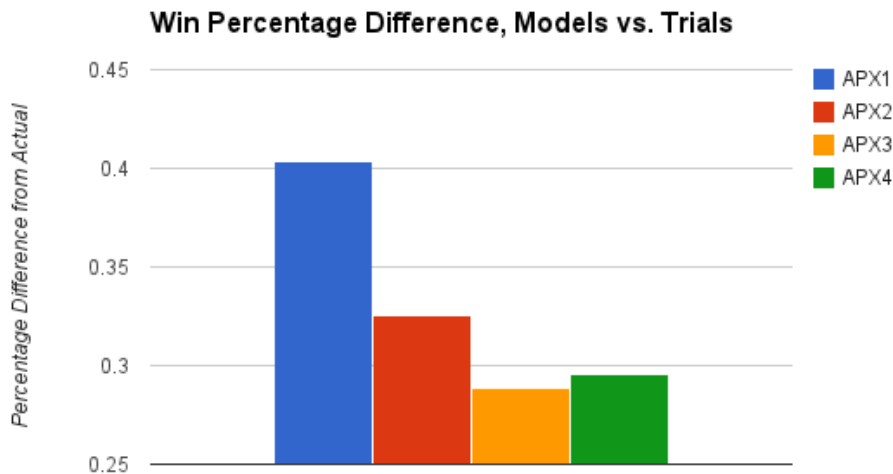

Figure 1: Win Percentage Difference, Models vs. Trials

Overall, the APX3 model turned out to be slightly more accurate than APX4 (Figure 1). As our goal was to create the most computationally efficient model that we could, it was reassuring to see that more complexity does not necessarily imply better results. Ideally, there may exist a model somewhere between APX3 and APX4 that minimizes error while keeping computational costs close to APX3.

The Terran-Zerg matches were the hardest to predict in our model. The Zerg basic melee unit, the Zergling, has relatively high damage (comparable to the more expensive Roach unit) that is offset by its need to actually be in melee range in order to attack. Because ranged units often "clump" into tightly clustered groups of units, this limits the surface area around which melee units can attack. Thus, applying the full DPS of all melee units to the opposing army may be an overly optimistic assessment. In addition, our random targeting algorithm means that we may attempt to target a unit with higher health and lower DPS, such as a Roach, instead of a more dangerous one that dies more quickly (e.g. a Zergling or Hydralisk). In addition, Ultralisks are very large units, and thus they are often slower than Zerglings when moving towards their targets. This greatly restricts the damage that they can do, although they can be devastating against groups of tightly clumped units, such as Marines, once they are within range.

Several additional factors can have a significant effect on the outcome of an engagement. Army positioning can affect which units are shot first, and poor positioning can change the outcome of an engagement. This is illustrated many times in our trial runs: in one particular case in the Round 3 Protoss-Zerg match, we see one case where Zerg won convincingly because the Protoss Archon was positioned in the front of the army and so was attacked first, where Zerg otherwise convincingly loses the match. Our model randomizes the order in which units are attacked, emulating "random" positioning; however, in a real engagement, players have much greater control over positioning, and so would almost certainly want to protect their high damage units by positioning them behind lower damage units with more health or armor.

## Conclusion

We created several different combat model for StarCraft 2 that approximate an engagement between two opposing armies. We constructed these models iteratively and assessed the impact on accuracy of simulating different aspects of the engagement. Our models focused primarily on the interaction between armies independent of map features, micromanagement, and unit positioning.

These approximation models have clear applications for any StarCraft 2 AI. Most StarCraft 2 AI systems have some type of combat model to determine whether or not it is reasonable to engage an opposing force. Our models can assist in this decision-making process by providing an estimated probability of a successful engagement, and also an average outcome of units. Additionally, it is useful for many AI systems to see what the impact of constructing a certain set of units might have on a future combat scenario. Here, too, our models provides utility at a computationally efficient rate.

## Limitations and Future Work

These combat models offer some insight into unit behavior in the general case; however, they are limited in that the results we produce reflect a random target selection, and the outcome of the models is based only on army composition, not unit position.

Our combat models also do not include spells, which can be extremely important in StarCraft 2. The most commonly used spells in StarCraft 2 are area-of-effect abilities capable of damaging or protecting large numbers of units and "zoning" abilities which make it more difficult for an opposing army to retreat or enemy melee units to engage.

There is also no notion of upgrades in our model, which affect the armor and damage of units. In many cases, early and mid-game attacks are timed with an upgrade advantage such that the attacking player has a higher level of upgrades than their opponent. The effect of these upgrades on the 1v1 unit level is well understood, but we have not extended this to our model, although it would be relatively straightforward to do so.

We provide no explicit modeling of air units in our combat model. Some ground units cannot target air units, and vise versa, so this may lead to some inaccuracy where our model is concerned. We did not evaluate the accuracy of our model in conflicts which include air units.

We can think of a number of ways to expand our models. We have already discussed spells above. A future combat model might be able to take some more sophisticated notions of positioning into account, which would allow an AI to enter into an engagement for which it would not be favored based on composition, but may be able to win because of current battlefield conditions. For example, the AI-controlled force may have higher ground or an important, high damage enemy unit may be exposed and vulnerable to sniping. In addition, StarCraft 2 is a game of imperfect information. Our model assumes that the AI player knows the exact composition of the opposing force, when in reality this is unlikely. A more sophisticated model might leverage a Partially Observable Markov Decision Process (POMDP) in order to guess at each army's chances of winning based only on observed enemy units and game time.

## Acknowledgements

The authors would like to thank Professor Amy Sliva for her guidance and feedback, Team Liquid for providing a website containing much of the unit information that we used, and all of the people at Blizzard who are responsible for making StarCraft 2.

## Resources

The code developed for this project, along with the StarCraft 2 maps and replay files, can be found at: https://bitbucket.org/karlwiegand/sc2apx.